\documentclass{article}

\usepackage{microtype}
\usepackage{graphicx}
\usepackage{subfigure}
\usepackage{booktabs} 

\usepackage{hyperref}

\usepackage[accepted]{icml2023}

\usepackage[utf8]{inputenc} 
\usepackage[T1]{fontenc}    
\usepackage{hyperref}       
\usepackage{url}            
\usepackage{booktabs}       
\usepackage{amsfonts}       
\usepackage{nicefrac}       
\usepackage{microtype}      
\usepackage{xcolor}         
\usepackage{amsmath}
\usepackage{amsthm}
\usepackage{booktabs}
\usepackage{algorithmic}
\usepackage{wrapfig,lipsum,booktabs}
\usepackage{tabularx}
\usepackage{graphicx}
\usepackage{amsmath}
\usepackage[T1]{fontenc}
\usepackage[utf8]{inputenc}
\usepackage{babel}
\usepackage[font=small,labelfont=bf]{caption}
\usepackage{algorithm}
\usepackage{algorithmic}
\usepackage{bm}
\usepackage{wrapfig}
\usepackage{lipsum}
\usepackage{xcolor}

\usepackage{blindtext}

\DeclareMathOperator*{\argmin}{arg\,min}

\newtheorem{theorem}{Theorem}
\newtheorem{corollary}{Corollary}[theorem]

 \renewcommand{\vec}[1]{\ensuremath{\mathbf{#1}}}

 \newcommand{\norm}[1]{\ensuremath{\| #1 \|}}

 \newcommand{\std}[1]{$\pm$ \tiny{#1}}

\newcommand{\blue}[1]{\textcolor{black}{#1}}

\icmltitlerunning{Wrapped Cauchy Distributed Angular Softmax for Long-Tailed Visual Recognition}

\begin{document}

\twocolumn[
\icmltitle{Wrapped Cauchy Distributed Angular Softmax for Long-Tailed Visual Recognition}



\icmlsetsymbol{equal}{*}

\begin{icmlauthorlist}
\icmlauthor{Boran Han}{yyy}
\end{icmlauthorlist}

\icmlaffiliation{yyy}{Amazon Web Services, AI. Work done while at Shell.}

\icmlcorrespondingauthor{Boran Han}{boranhan@amazon.com}

\icmlkeywords{Softmax, Wrapped Cauchy Distribution, Long-Tailed Visual Recognition}

\vskip 0.3in
]



\printAffiliationsAndNotice{}  

\begin{abstract}
Addressing imbalanced or long-tailed data is a major challenge in visual recognition tasks due to disparities between training and testing distributions and issues with data noise. We propose the Wrapped Cauchy Distributed Angular Softmax (WCDAS), a novel softmax function that incorporates data-wise Gaussian-based kernels into the angular correlation between feature representations and classifier weights, effectively mitigating noise and sparse sampling concerns. The class-wise distribution of angular representation becomes a sum of these kernels. Our theoretical analysis reveals that the wrapped Cauchy distribution excels the Gaussian distribution in approximating mixed distributions. Additionally, WCDAS uses trainable concentration parameters to dynamically adjust the compactness and margin of each class. Empirical results confirm label-aware behavior in these parameters and demonstrate WCDAS's superiority over other state-of-the-art softmax-based methods in handling long-tailed visual recognition across multiple benchmark datasets. The \href{https://github.com/boranhan/WCDAS_code}{code} is public available.
\end{abstract}

\section{Introduction}

Deep convolutional neural networks are the leading methods for computer vision tasks, including visual recognition. This strength is largely due to their robust representation learning, a technique that simplifies target images into a vector space with fewer dimensions. This crucial step is facilitated by the penultimate layer and subsequently fed into the final classifier, followed by a softmax function, which calculates the probability of an input image being in the $j$-th class: $P(y=j\mid \mathbf {x})$ \cite{10.5555/2969830.2969856, Goodfellow-et-al-2016}. However, most image recognition tasks have been demonstrated on well-balanced datasets. In contrast, most real-world data comes with an imbalanced distribution: a few high-frequency classes contain many training examples, while many low-frequency classes have insufficient training examples. This scenario is referred to as long-tailed recognition \cite{openlongtailrecognition}, and standard methods trained with such datasets tend not to yield the same performance as balanced ones \cite{openlongtailrecognition, DBLP:journals/corr/abs-1708-02002, cui2021parametric}. 

Numerous studies have focused on long-tailed recognition by attempting to re-balance the data distribution through class-balanced sampling or class re-weighting \cite{10.1007/11538059_91, kang2019decoupling, Kubt1997AddressingTC, 7780949, DBLP:journals/corr/abs-1806-00194, hong2023long}. However, they may under-represent the majority class \cite{10.1007/11538059_91, kang2019decoupling, Kubt1997AddressingTC} or destabilized the network during optimization \cite{7780949, DBLP:journals/corr/abs-1806-00194}. \blue{In addition to direct sampling, focal loss \cite{DBLP:journals/corr/abs-1708-02002} adopts loss function emphasizing  samples with larger loss value. However, it inevitably involves hyperparameters tuning by cross-validation. An alternative method is to adopt a label-aware correction via introducing a class-wise generalization error bound, such as Label-Distribution-Aware Margin Loss (LDAM) \cite{cao2019learning} and Balanced Meta-Softmax (BALMS) \cite{Ren2020balms}. Cao, \emph{et. al.} have proved that to improve the accuracy in recognizing long-tailed distributed data, classes with fewer training examples should have a higher generalization error bound \cite{cao2019learning}. However, both LDAM and BALMS can be vulnerable when the number of examples per class is unknown and constantly changing. Therefore, further corrections are required for continuous training. Meta-Weight-Net \cite{han2018coteaching} and Equilibrium loss \cite{Feng2021ExploringCE} are developed for class re-weighting and inter-class margin correction, which require no visibility to the underlying data distribution. However, those methods can either be subject to lengthy training time due to the nature of meta-learning \cite{han2018coteaching} or high space complexity because of the memory module \cite{Feng2021ExploringCE}. Lastly, using angular softmax, Kobayashi has proposed applying von Mises-Fisher distribution for compact feature space via a user-defined concentration parameter ($\kappa$) \cite{kobayashi2021cvpr}. However, such a method leads to lengthy hyper-parameter tuning with isotopic $\kappa$ for all classes. Meanwhile, their trainable class-wise $\bm{\kappa}$ approach shows inferior performance compared with the user-defined counterpart for an optimal performance \cite{kobayashi2021cvpr}.} In addition, data noise also exists in long-tail problem \cite{wu2021advlt, cao2020heteroskedastic, zhang2023deep}.

In light of these challenges, we propose the Wrapped Cauchy Distributed Angular Softmax (WCDAS) for long-tailed visual recognition \blue{based on \cite{kobayashi2021cvpr}}. We presume that the data-wise probability distribution follows the wrapped Normal distribution and deduce that WCDAS can be a better fit for mixed distributions comprised of individual distributions. We also demonstrate that WCDAS has several desirable features, such as adaptive regulation of the margins between classes via a concentration parameter, exhibiting label-aware behavior. Upon evaluation on several benchmark long-tailed image classification datasets, WCDAS outperforms state-of-the-art softmax-based methods. 

In summary, our contributions include: 1) proposing a model that considers noise-induced uncertainty in the form of data-wise wrapped Normal distributed kernels; 2) proving that WCDAS can more effectively fit the mixed distribution of such kernels; 3) showing that under a specific condition, our method also significantly enhances inter-class margins, resulting in compact clustering; and 4) demonstrating that the concentration parameter can be adaptive, with classes with fewer training samples having a higher concentration parameter and a larger margin.

\section{Related works}

\textbf{Angular-based Softmax.} Angular softmax \cite{liu2016large} and its mutant approaches \cite{8953658, Liu_2017_CVPR} have recently been proposed to improve the softmax loss in face verification tasks. Unlike conventional softmax, these methods allow neural networks to learn features in an angular manner by focusing on the cosine similarity between classifier weights and features.
Among these, Large-margin softmax \cite{liu2016large} directly enforces inter-class separability on the dot-product similarity, while SphereFace \cite{Liu_2017_CVPR} and ArcFace \cite{8953658} enforce multiplicative and additive angular margins on the hypersphere manifold, respectively. These margins are controlled by a hyperparameter, $m$: the larger the value of $m$, the larger the margin. Consequently, larger margins between classes can lead to compact clusters, resulting in enhanced performance over conventional softmax. \cite{Liu_2017_CVPR, 8953658, liu2016large}.

\textbf{Long-tailed recognition.} Datasets with long-tailed distribution \cite{openlongtailrecognition} not only have an imbalanced class with respect to the number of examples per class but also have a long tail of classes with only a few examples (<10), i. e., tail class. Two predominant approaches for such a problem are (1) loss function improvement and (2) data re-balancing. The former approach exploits aggressive learning in the tail classes \cite{DBLP:journals/corr/abs-1708-02002, tan2020eql, cui2019classbalancedloss} or forcing large margin between classes, especially tail classes\cite{cao2019learning, Ren2020balms, DBLP:journals/corr/abs-2001-01385}. In particular, Cao \emph{et. al.} \cite{cao2019learning} theoretically prove that the generalization error bound could be minimized by increasing the margins of tail classes. \blue{In addition to margin correction, Feng \emph{et. al.} also balances the classification via a Feature Memory Module \cite{Feng2021ExploringCE}.} At the same time, a handful of studies focus on data re-balancing during training, the second approach for imbalance training. Data rebalancing can be achieved by data re-sampling \cite{10.1007/11538059_91, kang2019decoupling, Kubt1997AddressingTC} or class re-weighting \cite{7780949, DBLP:journals/corr/abs-1806-00194}. However, data re-balancing-based strategies can lead to overfitting the tail classes and less efficient learning of the over-representative ones. \blue{The sampling strategies include fixed samplers \cite{kang2019decoupling} and meta-based samplers \cite{Ren2020balms, han2018coteaching}.} Decoupled training \cite{kang2019decoupling} is a simple yet effective solution that could significantly improve the generalization issue on long-tailed datasets. During this two-stage training, the representation learning is trained by instance-balanced sampler \cite{kang2019decoupling} while the classifier is further fine-tuned by class-balanced sampler \cite{kang2019decoupling} and meta sampler \cite{Ren2020balms}. 

\textbf{Parametric modeling of feature distribution.} Despite the emergence of deep learning being attributed to non-parametric non-linearity modeling, effectively training a network can prove challenging when dealing with certain real-world datasets that present issues such as class imbalance and insufficient examples. Parametric modeling, based on certain assumptions, can greatly assist learning in these adverse situations \cite{yang2021free, DBLP:journals/corr/abs-1901-07711}. One such approach involves approximating the Gaussian distribution of feature representation in few-shot learning to enhance generalizability \cite{yang2021free}. In the context of imbalanced classes, studies have shown that Gaussian distribution \cite{DBLP:journals/corr/abs-1901-07711} and von Mises-Fisher distribution \cite{kobayashi2021cvpr} modeling of feature representation, or angles between weights and features, can significantly improve performance. Parametric modeling of the feature space can also better handle uncertainty caused by noise in the data. Popular methods of utilizing parametric models to account for uncertainty include Variational Auto-encoder \cite{Kingma2014AutoEncodingVB}, Bayesian-based dropout \cite{pmlr-v48-gal16}, and DUL \cite{9157756}, among others.

Inspired by these three distinct approaches, we propose a method that parametrically models the feature representation. This method uses data-wise Gaussian kernels as basis and it includes class-wise parameters that are trainable, providing an adaptable framework for various types of data." 

\section{Wrapped Cauchy Distributed Angular Softmax (WCDAS)}

\textbf{Previous knowledge.} For angular softmax, the predicted probability from the linear classifier in CNNs for the $j$-th class given a sample vector $\vec x$ and a weighting vector $\vec w$ is formulated as:
\begin{equation}
\label{asoftmax}
P(y=j\mid \mathbf {\theta})= \frac{e^{f(\theta; j)}}{\sum _{c=1}^{C}e^{f(\theta; c)}}= {\frac {e^{s cos{\theta_j}}}{\sum _{c=1}^{C}e^{s cos{\theta_c}}}}
\end{equation}
where,
\begin{equation}
\label{fj}
f(\theta; j)= s \cos{\theta_j}
\end{equation}
$f(\theta; j)$ calculates the angle between normalized vectors $\mathbf {x}$ and $\mathbf {w}$, $cos{\theta_j} = \mathbf {x} ^{\mathsf {T}}\mathbf {w}_{j}$. For the ease of writing, we refer the angular representation ($\theta_j$) between $\mathbf {x}$ and $\mathbf {w}$ as "angular features". $s \in \mathbb{R}$ is a empirically-defined constant \cite{8953658, Liu_2017_CVPR} or trainable parameter \cite{kobayashi2021cvpr}. 

\textbf{Intuition and Overview of WCDAS.} The probability function ($f(\theta; j)$) of the angular softmax function (Equation \ref{fj}) describes the angle between representation features and classifier weights. As such, the classifier weights are optimized to minimize the loss function, given $cos{\theta_j}$. However, this approach may potentially lead to overfitting, especially when training with a few examples, as discussed in previous large-margin based cosine softmax studies \cite{kobayashi2021cvpr, liu2016large}, or with data containing noise, as reported by other studies \cite{wu2021advlt, cao2020heteroskedastic, zhang2023deep}. To address these issues, our method seeks an optimal parametric probability density function of $\theta_j$, conditioned on $y=j$, i.e., $P(\mathbf {\theta}\mid y=j)$. To achieve this, we initially propose using a data-wise Gaussian-based kernel as a basis. Intuitively, given $\theta$, such a kernel can model the data-wise uncertainty caused by the input noise or sparse sampling, instead of a direct class-wise distribution (Section \ref{s1}). By doing so, we can obtain the class-wise angular feature probability density function by summing the individual basis (Section \ref{s1}). Subsequently, we prove that this class-wise distribution can be more accurately approximated by a Wrapped Cauchy distribution, $f(\bm{\rho}, \theta; j)$, with a class-wise trainable concentration parameter, $\bm{\rho} \in \mathbb{R^C}$ (Section \ref{s2}). We provide insights into why our novel softmax is a better parametric distribution for representation feature modeling (Section \ref{s2}) and how it can create large margins under specific conditions (Section \ref{s3}). 

\subsection{Wrapped Normal Basis for Angular Feature Density Estimation.} \label{s1}

\textbf{Assumption.} To mitigate overfitting in the representation features, we approximate the uncertainty induced by noise or sparse sampling using a Gaussian distribution. Consequently, the angular feature of each data point follows the probability distribution of a Normal distribution in circular coordinates, i.e., a Wrapped Normal distribution or a von Mises-Fisher distribution. Given that the latter approximates the former distribution, we treat both distributions as equivalent for ease of discussion. This model of noise or sparse sampling-induced uncertainty using a Gaussian distribution has been widely utilized in various studies \cite{pmlr-v48-gal16, rasmussen2005gaussian, DBLP:journals/corr/abs-2011-06225}. Following this assumption, the probability distribution function of the angular feature for the $m$-th data point in the $j$-th class can be represented in the form of a Symmetric-Wrapped Stable (SWS) distribution \cite{Jammalamadaka2001Topics}: 
\begin{equation}
\label{h_1}
h(\rho, \theta; m, j) = \frac{1}{2\pi} \left(1+2\sum_{n=1}^{\infty} \rho_m^{n^a} \cos{n(\theta_m-\mu_m)} \right)
\end{equation}
where $n \in \mathbb{N}$, $\rho_m \in [0, 1)$ denotes concentration parameter of $m$-th data in $j$-th class, $\mu_m$ denotes the center of $j$-th class and $a \in (0, 2]$. When $a = 1$, Equation \ref{h_1} returns the wrapped Cauchy distribution and for $a = 2$, we get the wrapped Normal distribution \cite{Jammalamadaka2001Topics}. The bigger $\rho_m$ is, the more compact the wrapped Normal kernel is. 
Since $h(\rho, \theta; m, j)$ computes the probability $\theta_m$ belongs to $j$-th class with the optimized classifier weights, hence, for the correct class to be recognized based on Equation \ref{asoftmax}, $\mu_m \rightarrow 0$. 

\begin{figure}[t]
\centering
	\includegraphics[width = 8cm]{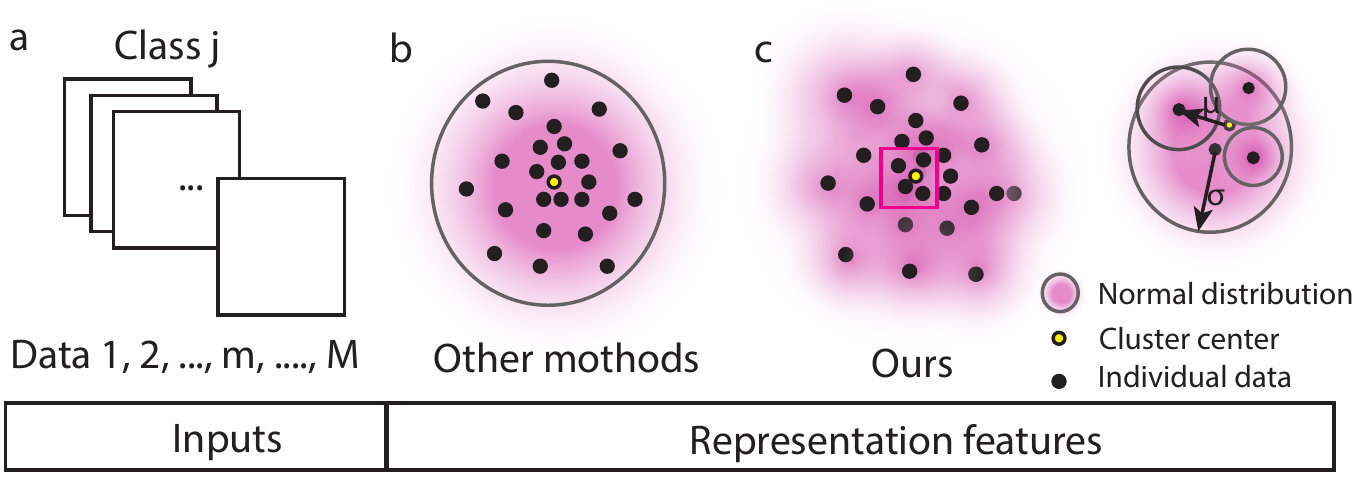}
\caption{\blue{Illustration of our method compared with other methods. Black dot: representation of each data in one class. Yellow dot with a black edge: centroid of the cluster. Gray solid line: Gaussian kernel boundary. (a) Input data 1, 2, ..., M in Class j. (b) parametric modeling of features from \emph{each data} via a wrapped Normal kernel. \cite{DBLP:journals/corr/abs-1901-07711, kobayashi2021cvpr} (c) Left panel: parametric modeling of features from \emph{each data} via wrapped Normal distribution. Right panel: zoomed diagram of the magenta box in the left panel.}}
	\label{WC_illustrator}
\end{figure}

\blue{Note that in our proposed method, we approximate the uncertainty of each $\theta_m$ as wrapped Normal distribution parameterized by $\rho_m$ and $\mu_m$ instead of modeling the $f(\theta; j)$ directly \cite{DBLP:journals/corr/abs-1901-07711, kobayashi2021cvpr}. Such difference is shown in Figure \ref{WC_illustrator}(b) and (c). }

\textbf{Class-wise probability distribution.} Subsequently, mixed distribution ${f(\theta; j)}$ can be obtained by summing all the $h(\rho, \theta; m, j)$ in $j$-th class:
\begin{equation}
\label{f_1}
f_\text{mixed}(\theta; j) = \frac{1}{M_j}\sum_{m=1}^{M_j} h(\rho, \theta; m, j) 
\end{equation}
where $M_j$ is the total number of samples in $j$-th class. $f_\text{mixed}(\theta; j)$ describes the mixture of $M_j$ wrapped Normal distributions centered around zero. Such an idea is used in the non-parametric estimation of a probability density function, such as kernel density estimation (\blue{KDE}) \cite{10.1214/aoms/1177728190, 10.2307/2237880}. However different from \blue{KDE}, $\bm{\rho}_j$, a vector comprised of all $\rho_m$ in $j$-th class, can be different in values, representing the heterogeneity of each data. 

\begin{theorem}
\label{t1}
Let $f_\text{mixed}(\theta; j)$ be a mixed distribution formed by summing several wrapped Normal distributions $h(\rho, \theta; m, j)$ (Equation \ref{h_1}). $h(\rho, \theta; m, j)$ is centered at $\mu_m$. $\mu_m$ follows Normal distribution $\mathcal{N}(0, \sigma)$ centered at zero, where $\sigma \rightarrow 0$. Then $f_\text{mixed}(\theta; j)$ can be approximated as:
\begin{equation}
\label{f_mixed}
f_\text{mixed}(\theta; j) \sim \frac{1}{2\pi M_j}\sum_{m=1}^{M_j} \left(1+2\sum_{n=1}^{\infty} \rho_m^{n^2} \cos{n\theta_m} \right)
\end{equation}
\end{theorem}

\begin{corollary}
\label{c1}
Let $f_\text{mixed}(\theta; j)$ be a mixed distribution formed by mixing several wrapped Normal distributions (Equation \ref{h_1} and Equation \ref{f_mixed}), then $f_\text{mixed}(\theta; j)$ is a wrapped distribution with cosine moments, $\alpha_\text{mixed}$, given by 
\begin{equation}
\alpha_\text{mixed}^{\{n\}} = \frac{1}{M_j}\sum_{m=1}^{M_j}\alpha_m^{\{n\}}
\end{equation}
where $\alpha_m^{\{n\}}$ is the n-th cosine trigonometric moment of $h(\rho, \theta; m, j)$. 
\end{corollary}

We present the detailed proof of Theorem \ref{t1} and Corollary \ref{c1} in \emph{Appendix A}. Theorem \ref{t1} essentially shows that when summing the wrapped Normal distributed kernel basis with a small perturbation away from zero, the result can be approximated as a sum of wrapped Normal distributions centered at zeros. Corollary \ref{c1} demonstrates that cosine moments of the mixed distribution can be obtained by averaging the cosine moments of each distribution. We note that the cosine moments of mixed distribution from two wrapped Normal distributions centered at zeros have been proven by Bailey, et. al. \cite{articleBailey}. We here prove that it can be generalized to several functions that are not centered at zero under certain conditions.

\subsection{Angular Feature Probability Approximation via Wrapped Cauchy Distribution.} \label{s2}
\begin{figure*}[h]
	\centering
	\includegraphics[width=1.0\textwidth]{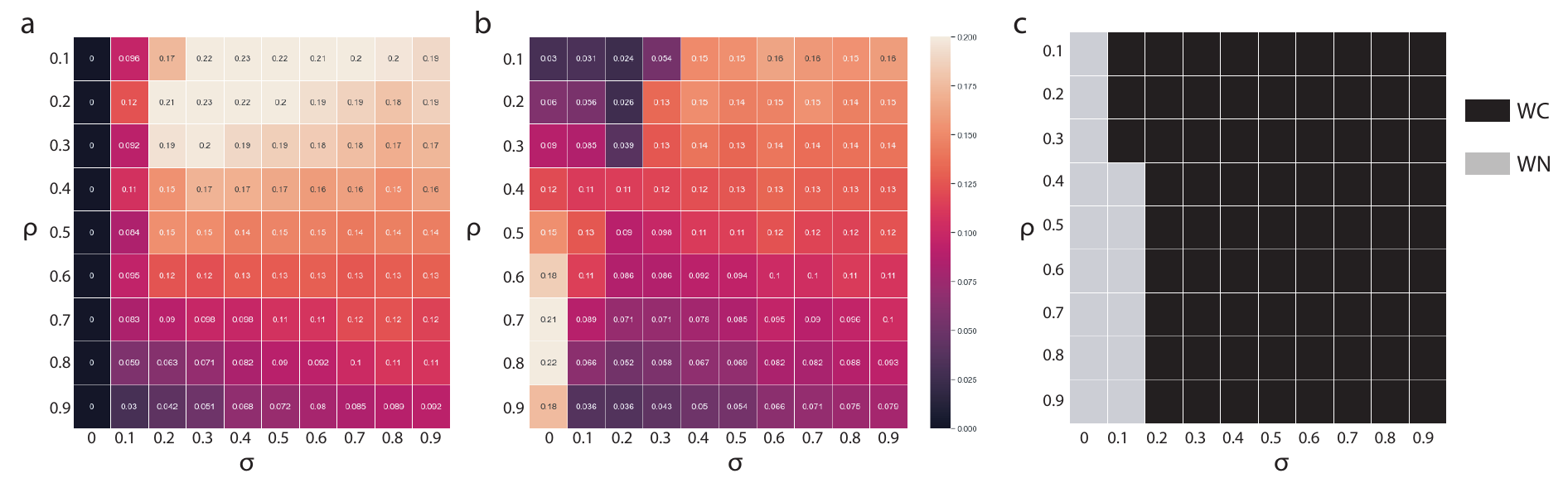}
\caption{Heatmap of $\varDelta_{\rho_\text{min}, \text {wn}}$ (a) and $\varDelta _{\rho_\text{min}, \text {wc}}$ (b) with respect to $\rho$ and $\sigma$. (c) Binary heatmap showing whether wrapped Cauchy (WC: black) or wrapped Normal (WN: gray) is preferred for simulated mixed distribution. }
	\label{simulation}
\end{figure*}

\blue{It is vital to find the optimal presentation of $f_\text{mixed}(\theta; j)$. One straightforward solution is to use non-parametric approaches \cite{10.1214/aoms/1177728190, 10.2307/2237880}. However, those methods usually require large computational costs for large dataset \cite{10.1006/jmva.1999.1863}. In our case, those methods also requires each $\rho_m$ to be calculated separately. Therefore, we approximate $f_\text{mixed}(\theta; j)$ with parametric distribution, denoting $f(\theta, \rho; j)$. According to Theorem \ref{t1}, \blue{$f(\theta, \rho; j)$} should also be an SWS distribution. Among the two predominant SWS distributions (Wrapped Cauchy distribution vs Wrapped Normal distribution), wrapped Cauchy distribution can fit Equation \ref{f_mixed} better than the wrapped Normal distribution.}

\begin{theorem}
\label{t1.1}
Let $f_\text{mixed}(\theta; j)$ be a mixed distribution formed by mixed several wrapped Normal distributions $h(\rho, \theta; m, j)$ of $j$-th class, centred around zero, defined in Equation \ref{f_1}. $f(\theta, \rho; j)$ is the approximated distribution with a choice of wrapped Normal $f_\text{WN}(\theta, \rho; j)$ and wrapped Cauchy $f_\text{WC}(\theta, \rho; j)$. Let $\rho_\text{j, min}$ of $f(\theta, \rho; j)$ minimize the least square error between $f(\theta, \rho; j)$ and mixed distribution $f_\text{mixed}(\theta; j)$ of $j$-th class: 
\begin{align} 
\label{delta}
&\varDelta_{\rho_j, \text{WC or WN}} =|| f_\text{WN or WC}(\rho, \theta; j) \nonumber \\
&- \frac{1}{2\pi M_j}\sum_{m=1}^{M_j} \left(1+2\sum_{n=1}^{\infty} \rho_m^{n^2} \cos{n\theta_m} \right) ||^2
\end{align}
\begin{align} 
\rho_\text{min}=\argmin_{\rho} \varDelta_\rho
\end{align}
Then the least square error of optimal $\rho_\text{j, min}$ of $j$-th class is correlated with standard deviation of $(\sum_{m_c=1}^{M_c} \rho_m^{n^2})^{\frac{1}{n}}$ or $(\sum_{m_c=1}^{M_c} \rho_m^{n^2})^{\frac{1}{n^2}}$ with respect to $n \in [1, \infty)$
\begin{align}
&\varDelta_{\rho_{j, \text{min}}, \text{WN}} \propto SD_{n = 1} \left(\sum_{m=1}^{M_j} \rho_m^{n^2} \right)^{\frac{1}{n}} \nonumber \\
&\text{or } \varDelta_{\rho_{j, \text{min}}, \text{WC}} \propto SD_{n=1} \left(\sum_{m=1}^{M_j} \rho_m^{n^2} \right)^{\frac{1}{n^2}} 
\end{align}
\end{theorem}

\begin{theorem}
\label{t2}
Let $\rho_m$ of individual $h(\rho, \theta; m, j)$ distribute uniformly across its defined domain $[0, 1)$, $\varDelta _{\text {WN}}$ and $\varDelta _{\text {WC}}$ defined in Equation \ref{delta}, Then $\varDelta_{\rho_{j, \text{min}}, \text{WC}} < \varDelta_{\rho_{j, \text{min}}, \text{WN}}$
\end{theorem}

The detailed proofs of Theorem \ref{t1.1} and Theorem \ref{t2} are provided in \emph{Appendix A}. These proofs substantiate that a mixed distribution constituted by SWS distributions aligns better with the wrapped Cauchy distribution than with the wrapped Normal distribution. Although the proof is analytical, it is based on the numerical assumption that $\rho_m$ in the $j$-th class is evenly distributed in $[0, 1)$ (Theorem \ref{t2}).

We also provide a numerical simulation for more general situations where $\rho_m \sim \mathcal{N}(\mu_{\rho}, \sigma_{\rho})$. Given that $\rho \in [0, 1)$, we simulate $\mu_{\rho}$ and $\sigma_{\rho}$ with the value of 0.1, 0.2, 0.3, 0.4, 0.5, 0.6, 0.7, 0.8, 0.9. Any $\rho$ values outside the $[0, 1)$ domain are clipped to 0 and 1, respectively. Figure \ref{simulation} shows the results, indicating a preference for the Wrapped Normal distribution when $\sigma_\rho$ is small ($\sigma_\rho \le 0.1$); otherwise, the wrapped Cauchy distribution is preferred. This implies that unless the mixed distribution $f_\text{mixed}(\theta; j)$ comprises wrapped Normal distributions with similar concentration parameters, the wrapped Cauchy distribution, due to its heavy tail, provides a better approximation for $f_\text{mixed}(\theta; j)$.

Assuming that $\rho_m$ follows a uniform distribution is an idealized assumption that simplifies the learning process. In practice, the actual distribution of $\rho_m$ might be more intricate. However, as our simulation of Gaussian-distributed $\rho_m$ demonstrates, there is a trend: the greater the diversity of $\rho_m$ values, the more advantageous the Cauchy distribution becomes as an approximation over Gaussian, given the heavy tail of the Cauchy distribution.


\subsection{Large Margin $\rho$ and optimization} \label{s3}

\begin{figure*}[t]
	\centering
	\includegraphics[width=0.8\textwidth]{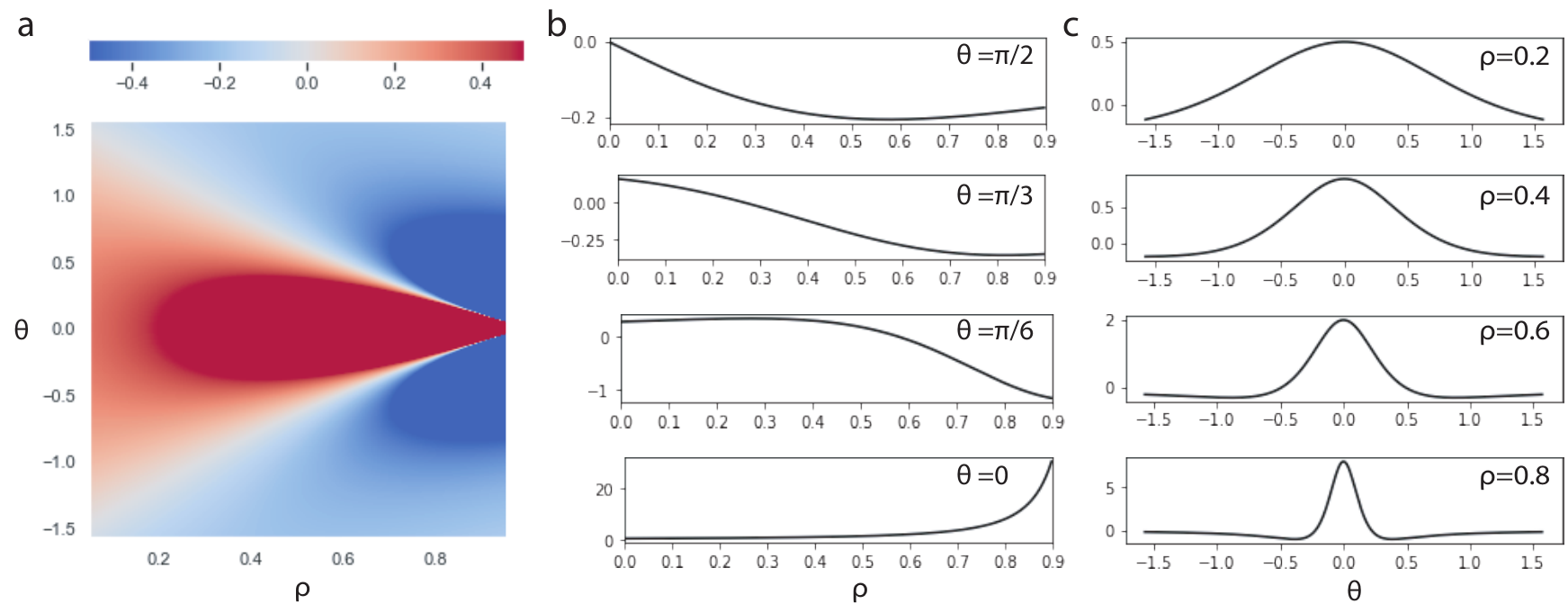}
\caption{Gradient plot of $\frac{\partial f(\bm{\rho}, \theta; j)}{\partial \bm{\rho}}$ with respect to $\rho$ and $\theta$(a). The cross sections plotted along $\rho$ (b) and $\theta$ (c).}
	\label{gradient}  
	\vspace{0mm}
\end{figure*}

It is important to obtain the optimal $\rho$. According to the geometric series, Equation \ref{h_1} can be written in an alternative form with element-wise calculation\cite{Jammalamadaka2001Topics}:
\begin{equation}
f(\bm{\rho}, \theta) = \frac{1-\bm{\rho}^2}{2\pi(1+\bm{\rho}^2-2\bm{\rho}\cos\theta))}
\end{equation}
where $\bm{\rho}$ is the vector containing $\rho_{j \in [1, C]}$ from all classes with the total number of $C$. \blue{We note that this alternative form of Equation \ref{h_1} is presented for the ease of calculating the margin between classes.}

\textbf{Large margin via WCDAS.} Several studies have demonstrated that the large margin-based softmax approach can lead to better performance both in balanced \cite{8953658, Liu_2017_CVPR, liu2016large} and imbalanced datasets \cite{Ren2020balms, DBLP:journals/corr/abs-1901-07711, cao2019learning}. We here prove that WCDAS can perform equivalently as those methods under a certain domain of $\bm{\rho}$. However, we note that not all $\bm{\rho}$ in WCDAS contribute to a large margin. Intuitively, only high $\bm{\rho}$ leads to tighter clustering. We here provide the boundary of $\bm{\rho}$ that will lead to large inter-class margins.

\begin{theorem}
\label{t3}
Let $\rho_j$ be the concentration parameter of wrapped Cauchy distribution $f_\text {wc}$ of the $j$-th class. $\vec x$ is the normalized presentation feature and $\vec w$ is the normalized weights of the classifier layer. Let $\theta_j$ and $\theta_k$ be the angle between $\vec x$ and $\vec w$ of $j$-th and $k$-th class respectively, where $\vec x$ is from class $j$. When $\rho_j \in (0.42332, 1)$, then $\norm{f_\text {WC}(\theta_j)-f_\text {WC}(\theta_k)} > \norm{\cos\theta_j-\cos\theta_k}$ for any $\theta_j$ and $\theta_k$ when $\cos\theta_j > \cos\theta_k$. The margin can be expressed as
\begin{equation}
\norm{f_\text {WC}(\theta_j)-f_\text {WC}(\theta_k)} = \frac{\rho_j+\rho_j^2}{\pi(1-\rho_j)^3} \norm {\cos\theta_j-\cos\theta_k}
\end{equation}
\end{theorem}

The detailed derivation is shown in \emph{Appendix A}. Theorem \ref{t3} shows that within such a domain, WCDAS yields a larger margin compared with $\cos \theta$. It is worth mentioning that such behavior holds with any $\theta_j$ and $\theta_k$. It is also shown that the larger $\rho_j$ is, the larger the margin is (\emph{Appendix Figure \ref{rho_curve}}). However, we note that our paper cannot prove that the margin is label-aware because of the gradient-based optimization. Therefore, the behaviors of $\rho$ during optimization require numerical studies (see Section \ref{ablation}). 

\textbf{Optimization} We can calculate its gradient with respect to $\rho$:
\begin{equation} \label{grd}
\frac{\partial f(\bm{\rho}, \theta)}{\partial \bm{\rho}} = \frac{-2 \bm{\rho} + (1 + \bm{\rho}^2) \cos\theta}{\pi (1 + \bm{\rho}^2 - 2 \bm{\rho} \cos\theta)^2}
\end{equation}
Through direct visualization of Equation \ref{grd} (Figure \ref{gradient}), we notice two characteristics of our method: (1) when $\theta$ is away from 0, $\rho$ decreases, i. e., $\frac{\partial f(\bm{\rho}, \theta)}{\partial \bm{\rho}}<0$; when $\theta$ is around 0, $\rho$ increases, i. e., $\frac{\partial f(\bm{\rho}, \theta)}{\partial \bm{\rho}}>0$. (2) The gradient $\frac{\partial f(\bm{\rho}, \theta)}{\partial \bm{\rho}}$ also increases when $\theta$ is around 0. Through the former characteristic, $\bm{\rho}$ is able to regulate the margin from the classifier layer. In contrast, the second characteristic can destabilize the whole network, since the value of $\bm{\rho}$ can also go beyond the defined domain. To address this issue, we define $\bm{w_{\rho}} \in (-\infty, \infty)$ so that $\bm{\rho}$ follows the behavior of sigmoid function with respect to $\bm{w_{\rho}}$, which approximates $\bm{\rho} \in [0, 1)$: 
\begin{equation}
\bm{\rho}=\frac{1}{1+e^{-\bm{w_{\rho}}}}, \bm{w_{\rho}} \in \mathbb{R^C}
\end{equation}

In summary, both the classifier and the feature extractor update the gradient. While the classifier is updated using our proposed method in Algorithm \ref{alg1}, the feature extractor (or encoder) is trained in a conventional manner.

\begin{algorithm}[h]
\caption{Wrapped Cauchy Distributed Angular Softmax}\label{alg1}
\begin{algorithmic}[1]
\STATE \textbf{Input:} Epoch number $E$, feature representation $\bm{x}$, weights in classifier $\vec{w}$, scale s. 
\STATE \textbf{Initialize:} $w_\rho$
\WHILE{$e < E$}
\WHILE{in Minibatch} 
\STATE $\bm{\rho}={\frac {1}{1+e^{-\bm{w}_{\rho}}}}$
\STATE $\cos\theta = \frac{\mathbf {x} ^{\mathsf {T}}\mathbf {w}_{j}}{\norm{\mathbf {x}}\norm{\mathbf {w}}}$
\STATE $f(\bm{\rho}, \theta) = \frac{1-\bm{\rho}^2}{2\pi(1+\bm{\rho}^2-2\bm{\rho}\cos\theta))}$
\STATE Compute Softmax: $\frac{e^{f(\rho, \theta; j)}}{\sum _{c=1}^{C}e^{f(\rho, \theta; c)}}$
\STATE Compute the cross entropy loss $L$
\STATE Update $\vec{w_\rho}, \vec{w}$ based on gradients $\frac{\partial L}{\partial \vec{w_\rho}}$, $\frac{\partial L}{\partial \vec{w}}$
\ENDWHILE
\STATE $ e \leftarrow e+1 $
\ENDWHILE
\end{algorithmic}
\end{algorithm}

\section{Empirical Experiments}

\subsection{Experimental setup} \label{setup}

We perform extensive ablation experiments on different aspects of our method (Section \ref{modeling} and \ref{ablation}). We also compared our approach with SOTA softmax-based methods (Section \ref{sota}) using four large-scale long-tailed datasets: CIFAR10-LT/100-LT \cite{krizhevsky2009learning}, ImageNet-LT \cite{openlongtailrecognition, 5206848} and iNaturalist 2018 \cite{DBLP:journals/corr/HornASSAPB17}. Among those datasets, CIFAR10-LT, CIFAR100-LT and ImageNet-LT are truncated from their balanced counterpart, following exponential decay across classes \cite{openlongtailrecognition} (see detail descriptions in \emph{Appendix B.1}). 

\textbf{Implementation.} All models are trained using SGD optimizer with momentum 0.9, weight decay $10^{-4}$. The learning rate decays by a cosine scheduler. Unless specified, we use 90 training epochs. Other hyper-parameters are listed in \emph{Appendix Table \ref{hp}}. The standard data augmentation is applied to input images. According to \cite{kang2019decoupling}, we apply a decoupled representation learning and classifier learning: The whole network is first trained via an instance-balanced sampler \cite{kang2019decoupling}. Only the classifier is further trained over 30 epochs sampled by a class-balanced sampler \cite{kang2019decoupling} or meta sampler \cite{Ren2020balms}. We apply WCDAS to both feature learning and classifier learning. 

\subsection{Wrapped Normal vs Wrapped Cauchy, Class-wise $\rho$ vs Single $\rho$ } \label{modeling}
\begin{table*}[h]
\centering
\begin{tabular}{lcccc|cccc}
\hline
$\rho$ & \multicolumn{4}{c|}{one $w_\rho$ for all classes ($w_\rho \in \mathbb{R}$)} & \multicolumn{4}{c}{class-wise $\bm{w_\rho}$ ($\bm{w_\rho} \in \mathbb{R^C}$)} \\
\hline
Method & Many & Medium & Few & All  & Many & Medium & Few & All \\
\hline
Angular Softmax & 52.8 & 33.9 & 15.7 & 38.7 & - & - & - & -\\
WNDAS & 55.0 & 38.2 & 20.4 & 42.1 & 54.9 & 38.6 & 20.2 & 42.3\\
WCDAS & \textbf{56.2} & \textbf{40.4} & \textbf{21.7} & \textbf{43.8} & \textbf{56.2} & \textbf{40.9} & \textbf{24.1} & \textbf{44.5} \\
\hline
\end{tabular}
\caption{Top 1 accuracy for ImageNet-LT (ResNet-10 \cite{https://doi.org/10.48550/arxiv.1512.03385}) with wrapped Normal distributed angular softmax (WNDAS) and WCDAS using one $w_\rho \in \mathbb{R}$ or class-wise $\bm{w_\rho} \in \mathbb{R^C}$. The result validates Theorem \ref{t2}}
\label{WNorWC}
\end{table*}

In this numerical experiment, we further validate Theorem \ref{t2} utilizing ImageNet-LT. For a fair comparison, Angular Softmax (Equation \ref{asoftmax}) is used as a baseline instead of the conventional softmax function. Note that we implement von Mises–Fisher distribution to approximate wrapped Normal distribution (WNDAS). Table \ref{WNorWC} shows that despite that both WNDAS and WCDAS display evident improvement from the baseline counterpart, WCDAS consistently performs better than WNDAS. Additionally, we test scenario when setting one $w_\rho$ for all classes ($w_\rho \in \mathbb{R}$) or class-wise $\bm{w_\rho}$ ($\bm{w_\rho} \in \mathbb{R^C}$). Our result proves that class-wise $\bm{w_\rho}$ shows superior performance. Intuitively, such results demonstrate that classes in the long-tailed training require different margins for better accuracy, consistent with previous observations \cite{cao2019learning, Ren2020balms}.

\subsection{$w_{\rho}$ optimization.} \label{ablation}

\begin{table}[h]
\centering
\begin{tabular}{lcccc}
    \hline
    Init. & Many & Medium & Few & All \\
    \hline
    2.0 & 55.6 & 40.5 & 23.2 & 43.9 \\
    1.0 & 57.3 & 40.5 & 21.4 & 44.3 \\
    0 & 56.0 & 40.7 & 23.5 & 44.2 \\
    He & 56.2  & 41.1  & 22.6  & 44.3 \\
    Xa. & 56.3 & 40.7  & 22.5 & 44.2 \\
    -1.0 & 56.2 & 40.9 & 24.1 & 44.5 \\
    -2.0 & 56.3 & 40.5 & 23.1 & 44.0 \\
    \hline
\end{tabular}
\caption{Top 1 accuracy for ImageNet-LT (ResNet-10 \cite{https://doi.org/10.48550/arxiv.1512.03385}) with various $w_{\rho}$ initialization (Init.) values. He \cite{7410480} and Xavier (Xa.) \cite{pmlr-v9-glorot10a}. Initial learning rate: 0.4}
\end{table}
\label{wi}

\textbf{Robustness of $w_{\rho}$ Initialization.} The initialization of parameters is a critical element in the optimization of deep networks, having significant impact on the quality of the final model. Given that our method introduces a new trainable parameter, $\bm{w_\rho}$, we performed empirical evaluations to assess its robustness under different initialization strategies. We observed some variance in the final outcomes depending on the initialization values used (\emph{Appendix Table \ref{wi_e150}}). This discrepancy, however, could be mitigated by either extending the number of training epochs (\emph{Appendix Table \ref{wi_e150}}) or increasing the learning rate (Table \ref{wi}). This suggests that shorter training periods or smaller learning rates may not be adequate for our approach. We also experimented with the He \cite{7410480} and Xavier \cite{pmlr-v9-glorot10a} initialization strategies, both of which are zero-centered. The results indicated that the final model was less sensitive to these initialization methods (Table \ref{wi}).

\begin{figure}[h]
    \centering
	\includegraphics[width=0.7\columnwidth]{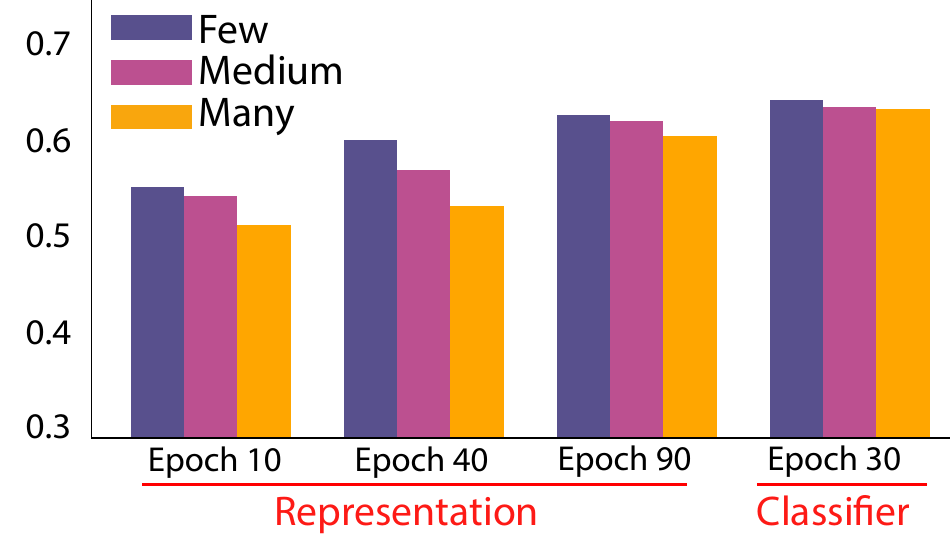}
    \captionof{figure}{Bar graph of $\rho$ with respect to three sets of class at different stages of training: 10th, 40th, 90th epoch at representation learning and 30th epoch at classifier learning. Three sets of class include few (<20), medium (20-100) and many (>100). Class-balanced sampler are used in classifier learning.}
	\label{weight per class}
\end{figure}

\begin{figure}[h]
    \centering
	\includegraphics[width=0.55\columnwidth]{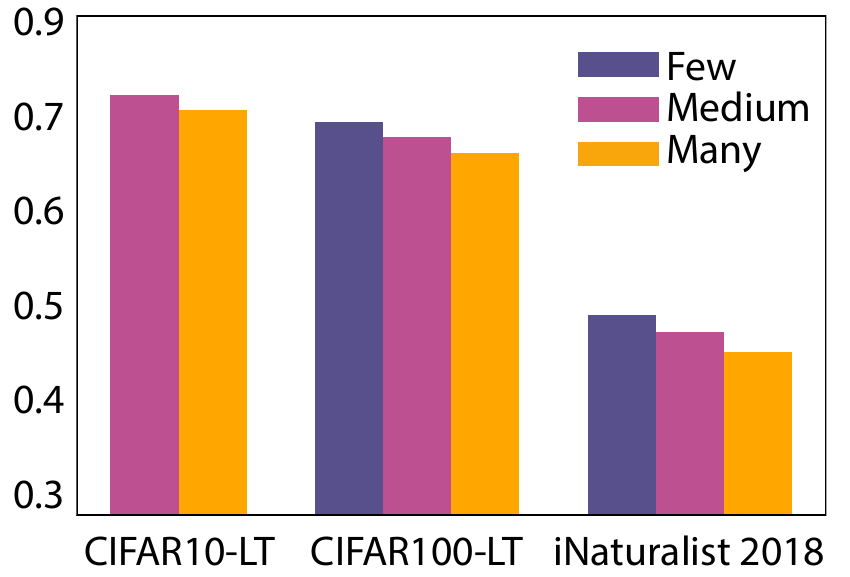}
\caption{Bar graph of $\rho$ values on CIFAR100-TL/10-TL and iNaturalist 2018.}
	\label{weight_dataset}
\end{figure}

\begin{table*}[h!]
\centering
\begin{tabular}{lccc|ccc}
\hline
Dataset & \multicolumn{3}{c|}{CIFAR-100-LT} & \multicolumn{3}{c}{CIFAR-10-LT} \\
\hline
Imbalance factor & 200 & 100 & 10 & 200 & 100 & 10 \\
\hline
Focal loss \cite{DBLP:journals/corr/abs-1708-02002} & 40.2 \std{0.5} & 43.8 \std{0.1} & 60.0 \std{0.6} & 71.8 \std{2.1} & 77.1 \std{0.2} & 90.3 \std{0.2} \\
LDAM loss \cite{cao2019learning} & 41.3 \std{0.4} & 46.1 \std{0.1} & 62.1 \std{0.3} & 73.6 \std{0.1} & 78.9 \std{0.9} & 90.3 \std{0.1}\\
cRT \cite{kang2019decoupling} & 44.5 \std{0.1} & 50.0 \std{0.2} & 63.3 \std{0.1} & 76.6 \std{0.2} & 82.0 \std{0.2} & 91.0 \std{0.0}\\
LWS \cite{kang2019decoupling} & 45.3 \std{0.1} & 50.5 \std{0.1} & 63.4 \std{0.1} & 78.1 \std{0.0} & 83.7 \std{0.0} & 91.1 \std{0.0}\\
BALMS \cite{Ren2020balms} & 45.5 \std{0.0} & 50.8 \std{0.0} & 63.0 \std{0.0} & 81.5 \std{0.0} & 84.9 \std{0.0} & 91.3 \std{0.0}\\
\hline
\multicolumn{3}{l}{Angular-based Softmax} \\
\hline
Angular Softmax & 44.2 \std {0.5} & 49.7 \std {0.6} & 64.1 \std {0.2} & 80.9 \std {0.2} & 83.8 \std {0.2} & 91.4 \std {0.1}\\
L-Softmax \cite{liu2016large} & 46.2 \std{0.2} & 51.3 \std{0.2} & 64.8 \std {0.1} & 79.9 \std{0.4} & 85.0 \std{0.2} & 91.8 \std{0.1}\\
AM-Softmax \cite{8953658} & 45.4 \std{0.4} & 50.1 \std{0.1} & 63.9 \std{0.2} & 77.5 \std{0.4} & 81.6 \std{0.5} & 90.9 \std{0.7}\\
t-vMF Similarity \cite{kobayashi2021cvpr} & 46.2 \std {0.2} & 50.3 \std {0.5} & 64.7 \std {0.2} & 80.9 \std {0.3} & 83.8 \std {0.3} & 91.2 \std {0.3}\\
\hline
WCDAS (ours) & \textbf{49.3} \std {0.1} & \textbf{52.5} \std {0.1}  & \textbf{65.8} \std {0.1} & \textbf{81.7} \std {0.1} & \textbf{86.4} \std {0.3} & \textbf{92.4} \std{0.2}\\
\hline
\end{tabular}
\caption{Top 1 accuracy (mean $\pm$ SD) for CIFAR-10/100-LT training with ResNet32 \cite{https://doi.org/10.48550/arxiv.1512.03385}. Results of Angular Softmax (Eq. \ref{asoftmax}), L-Softmax, AM-Softmax and t-vMF Similarity are reproduced with optimal hyper-parameters reported in their original papers. WCDAS generally outperforms SOTA methods.}
\label{cifartable}
\vspace{-0mm}
\end{table*}

\begin{table*}[h]
\centering
\begin{tabularx}{\textwidth}{lXXXX|XXXX}
\hline
Dataset & \multicolumn{4}{c|}{ImageNet-LT} & \multicolumn{4}{c}{iNaturalist 2018} \\
\hline
 & Many & Medium & Few & All & Many & Medium & Few & All \\
\hline
OLTR \cite{openlongtailrecognition} & 43.4 & 35.0 & 18.5 & 35.5 & 65.7 & 66.3 & 63.4 & 65.2 \\
Center loss \cite{10.1007/978-3-319-46478-7_31} & 53.0 & 35.1 & 15.6 & 39.1 & 71.7 & 66.0 & 60.4 & 64.3 \\
cRT \cite{kang2019decoupling} & 49.9 & 37.5 & 23.0 & 40.3 & 70.9 & 67.0 & 66.4 & 67.3 \\
LWS \cite{kang2019decoupling} & 48.0 & 37.5 & 22.9 & 39.6 & 69.0 & 68.2 & 66.6 & 67.7\\
BALMS \cite{Ren2020balms} & 48.0 & 38.3 & 22.9 & 39.9 & 66.8 & 67.4 & 67.9 & 68.1\\
\hline
\multicolumn{3}{l}{Angular-based Softmax}  \\
\hline
Angular Softmax (Eq. \ref{asoftmax}) & 52.8 & 33.9 & 15.7 & 38.7 & 71.8 & 65.3 & 61.4 & 65.0\\
L-Softmax \cite{liu2016large} & 54.0 & 35.1 & 15.4 & 39.1 & 72.7 & 66.1 & 60.1 & 64.5\\
AM-Softmax \cite{8953658} & 54.2 & 36.0 & 16.7 & 40.3 & 73.1 & 67.3 & 61.9 & 65.9\\
t-vMF Similarity \cite{kobayashi2021cvpr} & 55.4 & 39.9 & 22.5& 43.5 & 75.1 & \textbf{72.2} & 69.7 & 71.0 \\
\hline
WCDAS (class-balanced) & \textbf{56.2} & 40.9 & 24.1 & \textbf{44.5} & \textbf{75.5} & \textbf{72.3} & 69.8 & \textbf{71.8} \\
WCDAS (meta) &  53.8 & \textbf{41.7} & \textbf{25.3} & 44.1 & 71.4 & \textbf{72.3} & \textbf{70.5} & 70.8  \\
\hline
\end{tabularx}
\caption{Top 1 accuracy for ImageNet-LT (ResNet10 \cite{https://doi.org/10.48550/arxiv.1512.03385}) and iNaturalist 2018 (ResNet50 \cite{https://doi.org/10.48550/arxiv.1512.03385}). Results are reproduced with the same settings of our method (\emph{Appendix Table \ref{hp}}). Comparison of original results are provided in \emph{Appendix Table \ref{tableii}} together with more SOTA methods included. }
\label{tableiim}
\end{table*}

\textbf{Visualizing $\bm{\rho}$ During Optimization.} For a closer look at the optimization process, we graphically display the values of $\bm{\rho}$ during the two-stage decoupled learning phase, specifically for three class sets: few, medium, and many. With ImageNet-LT as an example (Figure \ref{weight per class}), we observe that $\bm{\rho}$ increases with each epoch, suggesting that the wrapped Cauchy distribution becomes increasingly tight. During representation learning, different class frequencies correspond to different values of $\rho$. \blue{On average, the 'Few' class exhibits a larger $\rho$ while the 'Many' class shows a smaller $\rho$ (\emph{Appendix Figure \ref{weight_init}}). Larger $\rho$ values lead to greater margins during training (Theorem \ref{t3}).} 

Prior research has established that both tighter feature clustering \cite{kobayashi2021cvpr} and larger margins \cite{cao2019learning, Ren2020balms} enhance classification results, especially for tail classes \cite{cao2019learning}. Our findings are consistent with these studies \cite{cao2019learning, kobayashi2021cvpr, Ren2020balms}. The frequency-dependent disparity in $\rho$ decreases in classifier learning due to the use of the class-balanced sampler \cite{kang2019decoupling}. It's also notable that class-dependent $\bm{\rho}$ values can be observed across all the tested datasets (Figure \ref{weight_dataset}). Moreover, our method shows that even with different initial positions, the aforementioned pattern holds true and $\bm{\rho}$ tends to converge to similar values (\emph{Appendix Figure \ref{weight_init}}), demonstrating stability during training.

\subsection{Comparing with SOTAs} \label{sota} 
We performed an extensive comparison of our method with state-of-the-art (SOTA) softmax-based methods designed for long-tail recognition on CIFAR-10/100-LT (Table \ref{cifartable}), ImageNet-LT (Table \ref{tableiim}), and iNaturalist 2018 (Table \ref{tableiim}). In addition, we included several leading angular-based softmax approaches for comparison, adhering to the same decoupled two-step training procedures. The class-balanced sampler was used for classifier learning in these methods. To ensure a fair comparison, our method also utilized the same sampler. A more detailed discussion about the choice of sampler is provided in \emph{Appendix B.4}. 

Given that WCDAS requires a larger learning rate (0.4) for ImageNet-LT and iNaturalist 2018, we sought to exclude the possibility that the superior results of our model could be attributed to the larger learning rate. To do this, we present two tables: one with SOTA methods reproduced using a learning rate of 0.4 (Table \ref{tableiim}), and the other featuring results directly obtained from the original papers (\emph{Appendix Table \ref{tableii}}). Both Table \ref{tableiim} and \emph{Appendix Table \ref{tableii}} indicate that our method achieves better accuracy than other competing softmax-based methods. 

Furthermore, the improvement is particularly noticeable in the tail class, which consists of fewer samples. For instance, our method improved the accuracy from 22.5 to 25.3 for ImageNet-LT, without compromising the head class. Previous works often sacrificed other classes in the process of improving accuracy \cite{kang2019decoupling, Ren2020balms}. This improvement was even more evident on CIFAR-100/10-LT, likely because fewer samples per class are more susceptible to noise, an aspect our method accounts for. 


\section{Conclusion} \label{conlusion}

We've introduced the WCDAS approach for long-tail visual recognition tasks. Generally, WCDAS outperforms state-of-the-art (SOTA) softmax-based methods across all four datasets. The symmetric-wrapped stable (SWS) family incorporates a wide variety of distributions, each with its unique properties \cite{Jammalamadaka2001Topics}. Our work expands the understanding of their utility in various contexts and challenges established methods, such as vMF. \blue{Four distinct \emph{advantages} distinguish our method from previous works \cite{kobayashi2021cvpr, cao2019learning} and contribute to its superior performance: (1) WCDAS accommodates out-of-distribution "imperfect" data due to its heavy tail, while still ensuring compact intra-class feature clustering. (2) WCDAS operates like a large margin angular softmax when $\rho$ is large. As $\rho$ increases during training, our loss function aligns with the classification task's cross-entropy loss. We provide a visualization of the loss surface with respect to $\rho$ and $\theta$. (3) Our empirical study shows that tail classes have larger $\rho$, leading to more compact clusters and larger margins (Theorem \ref{t3}). Previous studies have confirmed the significant performance benefits of these factors \cite{cao2019learning}. (4) Our method, unlike previous user-defined parameter approaches \cite{kobayashi2021cvpr}, achieves optimal performance with trainable $\rho$.}

However, WCDAS has \emph{limitations}: it may necessitate a different learning rate or number of epochs compared to other methods, indicating a need for parameter re-tuning. Although WCDAS displays label-aware behaviors for all tested datasets, our paper does not offer a theoretical proof for this. 

Looking to the future, WCDAS can serve as the softmax function replacement in deep learning models, improving other deep learning methods, such as mixture-of-experts \cite{wang2020long, zhang2021test}, and contrastive learning-based methods \cite{cui2021parametric, https://doi.org/10.48550/arxiv.2103.14267}. Additionally, WCDAS could potentially be applicable to long-tail video recognition \cite{Zhang_2021_ICCV} and long-tail object detection \cite{Feng2021ExploringCE} with minimal or no adjustments. However, further validation is necessary for these domains.

\bibliographystyle{main}  
\bibliography{main}

\onecolumn
\appendix
\section{Proofs and Derivations}

\subsection{Proof to Theorem 1} \label{theorem 1}

\begin{equation}
\label{h_1}
h(\rho, \theta; m, j) = \frac{1}{2\pi} \left( 1+2\sum_{n=1}^{\infty} \rho_m^{n^2} \cos{n(\theta_m-\mu_m)} \right), n \in \mathbb{N}
\end{equation}

\begin{equation}
\label{h_1}
h(\rho, \theta; m, j) = \frac{1}{2\pi} \left( 1+2\sum_{n=1}^{\infty} \rho_m^{n^2} (\sin{n\mu_m}\sin{n\theta_m}+\cos{n\mu_m}\cos{n\theta_m}) \right)
\end{equation}

Given that $\mu_m \rightarrow 0$, then $\sin{n\mu_m}$ can be approximated as $n\mu_m$ and $\cos{n\mu_m}$ can be approximated as 1. Therefore:

\begin{equation}
\label{h_1}
h(\rho, \theta; m, j) = \frac{1}{2\pi} \left(1+2\sum_{n=1}^{\infty} \rho_m^{n^2} (n\mu_j\sin{n\theta_m}+\cos{n\theta_m}) \right)
\end{equation}

Therefore, mixed distribution $f_\text{mixed}(\theta; j)$ can be written as:

\begin{equation}
\tiny
\label{h_1}
f_\text{mixed}(\theta; j) = \frac{1}{M_j} \left(\underbrace{\frac{1}{2\pi} \left(1+2\sum_{n=1}^{\infty} \rho_1^{n^2} (n\mu_1\sin{n\theta_1}+\cos{n\theta_1}) \right) + \dots + \frac{1}{2\pi} \left(1+2\sum_{n=1}^{\infty} \rho_{M_j}^{n^2} (n\mu_{M_j}\sin{n\theta_{M_j}}+\cos{n\theta_{M_j}}) \right)}_{M_j} \right)
\end{equation}

\begin{align}
\label{h_1}
f_\text{mixed}(\theta; j) &= \frac{1}{2\pi} + \frac{1}{M_j}\underbrace{\left(2\sum_{n=1}^{\infty}  (n\rho_1^{n^2}\mu_1\sin{n\theta_1} + \dots + n\rho_{M_j}^{n^2}\mu_{M_j}\sin{n\theta_{M_j}}) \right)}_{M_j} \\&+ \frac{1}{M_j}\underbrace{\left(2\sum_{n=1}^{\infty} (\rho_1^{n^2}\cos{n\theta_1}+ \dots+\rho_{M_j}^{n^2}\cos{n\theta_{M_j}})\right)}_{M_j}
\end{align}

Given that $\mu$ follows $\mathcal{N}(0, \sigma)$, Therefore we can further approximate:

\begin{equation}
\frac{1}{M_j}\underbrace{\left(2\sum_{n=1}^{\infty}  (n\rho_1^{n^2}\mu_1\sin{n\theta_1} + \dots + n\rho_{M_j}^{n^2}\mu_{M_j}\sin{n\theta_{M_j}}) \right)}_{M_j} \rightarrow 0
\end{equation}

Subsequently, we obtain: 

\begin{equation}
f_\text{mixed}(\theta; j) = \frac{1}{2\pi M_j}\sum_{m=1}^{M_j} \left(1+2\sum_{n=1}^{\infty} \rho_m^{n^2} \cos{n\theta_m} \right)
\end{equation}

\subsection{Proof to Corollary 1.1}

\begin{align}
f_\text{mixed}(\theta; j) &= \frac{1}{2\pi M_j}\sum_{m=1}^{M_j} \left(1+2\sum_{n=1}^{\infty} \rho_m^{n^2} \cos{n\theta_m} \right) \\
&= \frac{1}{2\pi} \left(1+2\sum_{n=1}^{\infty} \frac{1}{M_j}\sum_{m=1}^{M_j} \rho_m^{n^2}  \cos{n\theta_m} \right)
\end{align}

Therefore, we get: 

\begin{align}
f_\text{mixed}(\theta; j) &= \frac{1}{2\pi} \left(1+2\sum_{n=1}^{\infty} \alpha_\text{mixed}^{\{n\}} \cos{n\theta_m} \right) \\
\end{align}
where $\alpha_\text{mixed}$ is the cosine moment of the mixed distribution:

\begin{equation}
\alpha_\text{mixed}^{\{n\}} = \frac{1}{M_j}\sum_{m=1}^{M_j} \rho_m^{n^2} = \frac{1}{M_j}\sum_{m=1}^{M_j}\alpha_m^{\{n\}}
\end{equation}

\subsection{Proof to Theorem 2}

\begin{align} 
\varDelta_\rho&=\frac{1}{\pi}\sum_{n=1}^\infty \left( (\alpha_{\text{fit}}^{\{n\}}-\alpha_\text{mixed}^{\{n\}}) \cos{n\theta}\right)^2
\end{align}

For any $\theta$ and $n$, to minimize $\varDelta_\rho$, it is equivalently as minimizing $\sum_{n=1}^\infty (\alpha_{\text{fit}}^{\{n\}}-\alpha_\text{mixed}^{\{n\}})^2$:
\begin{align} 
\varDelta_\rho = \frac{1}{\pi}\sum_{n=1}^\infty (\alpha_{\text{fit}}^{\{n\}}-\alpha_\text{mixed}^{\{n\}})^2 
\end{align}

For wrapped Cauchy distribution, $\alpha_{\text{fit}}^{\{n\}} = \rho_\text{WC}^n, \forall \rho \in \mathbb{R}$. For wrapped Normal distribution, $\alpha_{\text{fit}}^{\{n\}} = \rho_\text{WN}^{n^2}, \forall \rho \in \mathbb{R}$. $\alpha_\text{mixed}^{\{n\}} = \frac{1}{M_j}\sum_{m=1}^{M_j} \rho_m^{n^2} $ according to Corollary 1.1. Without losing the generosity, we derive the case of wrapped Cauchy distribution as an example:
\begin{align} 
\label{delta}
\varDelta_{\rho_\text{min}, \text {WC}} = \frac{1}{\pi}\sum_{n=1}^\infty (\rho^n_\text{min}- \frac{1}{M_j}\sum_{m=1}^{M_j} \rho_m^{n^2})^2 = \frac{1}{\pi}\sum_{n=1}^\infty (\rho^n_\text{min}- \alpha_\text{mixed}^{\{n\}})^2 
\end{align}

$(\alpha_\text{mixed}^{\{n\}})^\frac{1}{n} = (\frac{1}{M_j}\sum_{m=1}^{M_j} \rho_m^{n^2}))^\frac{1}{n}$ can be treated as the $n$-th component of a cluster. To minimize the error $\varDelta_{\rho_\text{min}, \text {wc}}$, $\rho_\text{min}$ is the centroid of the cluster composed of $n$ number of $ (\frac{1}{M_j}\sum_{m=1}^{M_j} \rho_m^{n^2})^\frac{1}{n}$. Therefore,

\begin{align} 
\label{rho_min}
\rho_\text{min}=\mathbb{E}_{n \in [1, \infty)}(\frac{1}{M_j}\sum_{m=1}^{M_j} \rho_m^{n^2})^\frac{1}{n} = \mathbb{E}_{n \in [1, \infty)}(\alpha_\text{mixed}^{\{n\}})^\frac{1}{n}
\end{align} 

Take Equation \ref{rho_min} back into Equation \ref{delta} substituting $\rho_\text{min}^n$, we approximate:

\begin{align} 
\label{123}
\varDelta_{\rho_\text{min}, \text {wc}} &=\frac{1}{\pi} \sum_{n=1}^\infty \left((\mathbb{E}_{n \in [1, \infty)}(\alpha_\text{mixed}^{\{n\}})^\frac{1}{n})^n- \alpha_\text{mixed}^{\{n\}}\right)^2 \nonumber \\
&= \frac{1}{\pi} \sum_{n=1}^\infty \left((\mathbb{E}_{n \in [1, \infty)}(\alpha_\text{mixed}^{\{n\}})^\frac{1}{n})^n- ((\alpha_\text{mixed}^{\{n\}})^\frac{1}{n})^n\right)^2
\end{align}
According to Binomial Theorem:
\begin{align}
\label{bio}
x^n-a^n &=(x-a)(a^{n-1}+ x a^{n-2} +\cdots x^{n-2}a + x^{n-1}).
\end{align}

Let $\overline{a} = \mathbb{E}_{n \in [1, \infty)}(\alpha_\text{mixed}^{\{n\}})^\frac{1}{n}$ and $a_n = (\alpha_\text{mixed}^{\{n\}})^\frac{1}{n}$, we simplify Equation \ref{123} into:

\begin{align}
\label{bio}
\varDelta_{\rho_\text{min}, \text {WC}} &\propto  \sum_{n=0}^\infty (\overline{a}^n-a_n^n)^2 \nonumber\\ &\propto (\overline{a}-a_1)^2 +(\overline{a}-a_2)^2(\overline{a}+a_2)^2 + \cdots \nonumber \\&+ (\overline{a}-a_{n-1})^2(a^{n-1}+ \overline{a} a^{n-2} +\cdots \overline{a}^{n-2}a + \overline{a}^{n-1})^2.
\end{align}

We expand the Equation \ref{123} based on Equation \ref{bio}:
\begin{align} 
\label{321}
\varDelta_{\rho_\text{min}, \text {WC}} &= \frac{1}{\pi} \sum_{n=1}^\infty \left(\mathbb{E}_{n \in [1, \infty)}(\alpha_\text{mixed}^{\{n\}})^\frac{1}{n}- (\alpha_\text{mixed}^{\{n\}})^\frac{1}{n}\right)^2 \left((\mathbb{E}_{n \in [1, \infty)}(\alpha_\text{mixed}^{\{n\}})^\frac{1}{n})^{n-1} +\cdots + (\alpha_\text{mixed}^{\{n\}})^\frac{n-1}{n}\right)^2
\end{align}

Because $\rho_m \in [0, 1)$, $\alpha_\text{mixed}^{\{n\}} \in [0, 1)$ and the value of $\alpha_\text{mixed}^{\{n\}}$ decrease as $n$ increases. Therefore, higher order terms in Equation \ref{321} can be neglected ($n$>1). Accordingly, we get: 

\begin{align}
\label{delta_wc}
\varDelta_{\rho_\text{min}, \text {WC}} &\sim \frac{1}{\pi} \left(\mathbb{E}_{n \in [1, \infty)}(\alpha_\text{mixed}^{\{n\}})^\frac{1}{n}- (\alpha_\text{mixed}^{\{1\}})\right)^2 + \mathcal{O}(n) \nonumber\\
&\sim \frac{1}{\pi} \left(\mathbb{E}_{n \in [1, \infty)}(\alpha_\text{mixed}^{\{n\}})^\frac{1}{n}- \frac{1}{M_j}\sum_{m=1}^{M_j} \rho_m \right)^2 + \mathcal{O}(n) \nonumber\\
&\propto SD_{n = 1}(\frac{1}{M_j}\sum_{m=1}^{M_j} \rho_m^{n^2})^{\frac{1}{n}}
\end{align}

Following a similar derivation, when $\alpha_{\text{fit}}^{\{n\}} = \rho^{n^2}$: 
\begin{align} 
\label{delta_wn}
\varDelta_{\rho_\text{min}, \text {WN}} &\sim \frac{1}{\pi} \left(\mathbb{E}_{n \in [1, \infty)}(\alpha_\text{mixed}^{\{n\}})^\frac{1}{n^2}- (\alpha_\text{mixed}^{\{1\}})\right)^2 + \mathcal{O}(n) \nonumber\\
&\sim \frac{1}{\pi} \left(\mathbb{E}_{n \in [1, \infty)}(\alpha_\text{mixed}^{\{n\}})^\frac{1}{n^2}- \frac{1}{M_j}\sum_{m=1}^{M_j} \rho_m \right)^2 + \mathcal{O}(n) \nonumber\\
&\propto SD_{n=1}(\frac{1}{M_j}\sum_{m=1}^{M_j} \rho_m^{n^2})^{\frac{1}{n^2}}
\end{align}

\subsection{Proof to Theorem 3}

Let $\rho_m$ of individual $h(\rho, \theta; m, j)$ distribute uniformly across its defined domain $[0, 1)$. Assuming that we have $N$ number of $h(\rho, \theta; m, j)$ (i. e., $M_j = N$), then $\rho_1 = 1/N$,  $\rho_1 = 2/N$, $\cdots$, $\rho_N = (N-1)/N$.

\begin{equation}
\alpha_\text{mixed}^{\{n\}} = \frac{1}{N}\sum_{m=1}^{M} \left(\frac{m}{N}\right)^{n^2} 
\end{equation}

Given Faulhaber's formula, which is: 

\begin{align} 
\sum _{k=1}^{N}k^{p}={\frac {N^{p+1}}{p+1}}+{\frac {1}{2}}N^{p}+\sum _{k=2}^{p}{\frac {B_{k}}{k!}}\frac{p!}{(p-k+1)!}N^{p-k+1}
\end{align}

The coefficients involve Bernoulli numbers $B_j$. For each $n$, we get:

For $n=1$, 
\begin{align} 
\label{sum_1}
\alpha_\text{mixed}^{\{1\}} = \frac{1}{N} \sum_{k=1}^{N} \rho = \frac{1}{N} \sum_{k=1}^{N} \frac{k}{N} =(1 + \frac{1}{N}) / 2 
\end{align}

For $n=2$, 
\begin{align} 
\alpha_\text{mixed}^{\{2\}} = \frac{1}{N}\sum_{k=1}^{N}\rho^4 = \frac{1}{N} \sum_{k=1}^{N} \left(\frac{k}{N}\right)^4 = \frac{1}{5} + \frac{1}{2}\frac{1}{N} + \frac{1}{3}\frac{1}{N^2} - \frac{1}{30} \frac{1}{N^4}
\end{align}

For $n=3$, 
\begin{align} 
\alpha_\text{mixed}^{\{3\}} = \frac{1}{N}\sum_{k=1}^{N}\rho^9 = \frac{1}{N} \sum_{k=1}^{N} \left(\frac{k}{N}\right)^9 = & \frac{1}{10} + \frac{1}{2}\frac{1}{N} + \frac{3}{4}\frac{1}{N^2} - \frac{7}{10}\frac{1}{N^4} + \frac{1}{2}\frac{1}{N^6} - \frac{3}{20}\frac{1}{N^8}\\
&\dots
\end{align}

For $n=n_0$, 
\begin{align} 
\label{n_sum}
\alpha_\text{mixed}^{\{n\}} = \frac{1}{N}\sum_{k=1}^{N}\rho^{n_0^2} = \frac{1}{N} \sum_{k=1}^{N} \left(\frac{k}{N}\right)^{n_0^2} = {\frac {1}{{n_0^2} +1}}+{\frac {1}{2N}}+\frac {1}{N}\sum _{k=2}^{n_0^2}{\frac {B_{k}}{k!}}\frac{n_0^2!}{(n_0^2-k+1)!}\frac{1}{N^{k-1}}
\end{align}

Given $N>1$, then $\alpha_\text{mixed}^{\{n\}} \in (0, 1)$ (Equation \ref{sum_1} - \ref{n_sum}). Hence, $(\alpha_\text{mixed}^{\{n\}})^\frac{1}{n}$ shows more "variance" than $(\alpha_\text{mixed}^{\{n\}})^\frac{1}{n^2}$ given $\alpha_\text{mixed}^{\{n\}} \in (0, 1)$. Therefore, $(\mathbb{E}_{n \in [1, \infty)}(\alpha_\text{mixed}^{\{n\}})^\frac{1}{n} - \alpha_\text{mixed}^{\{1\}})^2 < (\mathbb{E}_{n \in [1, \infty)}(\alpha_\text{mixed}^{\{n\}})^\frac{1}{n^2} - \alpha_\text{mixed}^{\{1\}})^2$. According to Equation \ref{delta_wc} and Equation \ref{delta_wn}:
\begin{align} 
\varDelta_{\rho_\text{min}, \text {WC}} < \varDelta_{\rho_\text{min}, \text {WN}}
\end{align}

\subsection{Proof to Theorem 4}

\begin{align}
\norm{f_\text {wc}(\theta_1)-f_\text {wc}(\theta_2)} =&\norm{\frac{1-\rho^2}{2\pi(1+\rho^2-2\rho\cos\theta_1)}-\frac{1-\rho^2}{2\pi(1+\rho^2-2\rho\cos\theta_2)}} \\
&= \frac{1-\rho^2}{2\pi} \norm{\frac{2\rho \cos\theta_1-\cos\theta_2}{(1+\rho^2-2\rho\cos\theta_1)(1+\rho^2-2\rho\cos\theta_2)}} \\
&\ge \frac{1-\rho^2}{2\pi} \frac{2\rho \norm{\cos\theta_1-\cos\theta_2}}{(1+\rho^2-2\rho)(1+\rho^2-2\rho)}
\end{align}
Simplified the above equation as follows:
\begin{align}
&\frac{1-\rho^2}{2\pi} \frac{2\rho \norm{\cos\theta_1-\cos\theta_2}}{(1+\rho^2-2\rho)(1+\rho^2-2\rho)} \\
= &\frac{\rho+\rho^2}{\pi(1-\rho)^3} \norm{\cos\theta_1-\cos\theta_2}
\end{align}
In order to show a larger margin, it needs to satisfy the following condition:
\begin{align}
\frac{\rho+\rho^2}{\pi(1-\rho)^3} \norm{\cos\theta_1-\cos\theta_2} &\ge \norm{\cos\theta_1-\cos\theta_2} \\
\label{rho_solver}
\frac{\rho+\rho^2}{\pi(1-\rho)^3} &\ge 1
\end{align}

\begin{figure}[h]
	\centering
	\includegraphics[width= 0.7\textwidth]{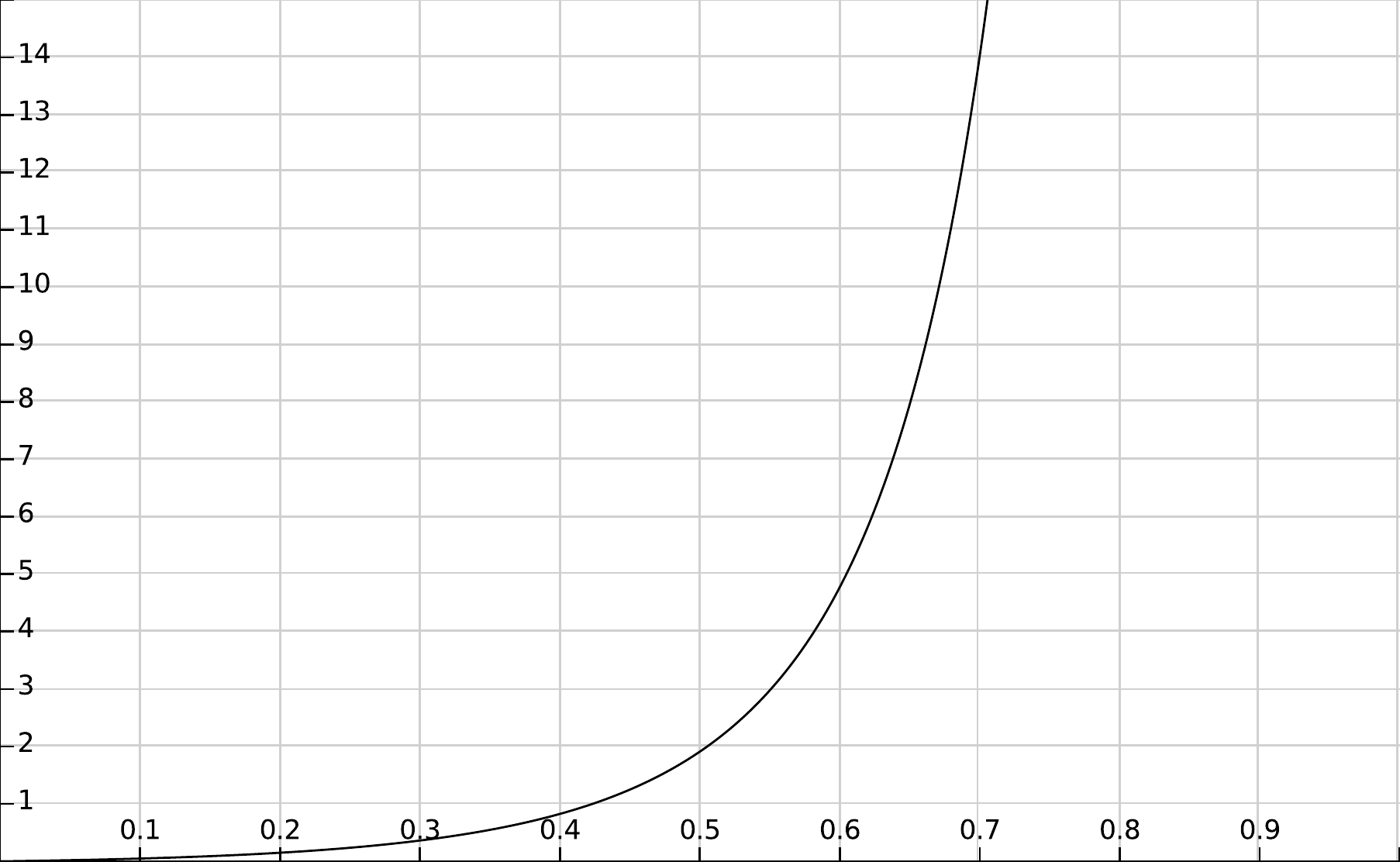}
\caption{plot of $\frac{\rho+\rho^2}{\pi(1-\rho)^3}$ (Y-axis) with respect to $\rho$ (X-axis). }
	\label{rho_curve}
\end{figure}

Solving Equation \ref{rho_solver}, we get $\rho \ge 0.42332$. Additionally, notice from Equation \ref{rho_solver} that the larger $\rho$ is, the larger the margin is (Figure \ref{rho_curve}). 

\section{Supplementary Results}
\subsection{Experiment settings}

\textbf{CIFAR10-LT and CIFAR100-LT}: CIFAR10-LT and CIFAR100-LT contain 10 and 100 classes, respectively. Various imbalance factors (10-200) are evaluated. an imbalance factor $\beta$ is calculated by $\beta = \frac{M_\text{max}}{M_\text{min}}$ where $M_\text{max}$ and $M_\text{min}$ are the numbers of training samples for the most and least frequent classes respectively. We employ the ResNet-32 backbone for these two datasets, similar to previous works. Given that CIFAR-LT 10/100 tends to show large variances in performance results, as stated in \cite{Ren2020balms}, we, therefore, report the mean and standard error from 3 independent replicas.

\textbf{ImageNet-LT}: It contains 1000 classes, and the number of images per class ranges from 1280 to 5 images with an imbalance factor of 256. ResNet-10 and \blue{ResNext-50 backbones are} used for the experiments. ImageNet-LT is also used for various ablation studies. 

\textbf{iNaturalist 2018}: It is a naturally imbalanced fine-grained dataset with 8,142 categories, following the long-tailed distribution. The number of images per class ranges from 1000 to 2, with an imbalance factor of 500. We use ResNet-50 as the backbone and apply the same training settings as for ImageNet-LT except batch size 512.

\textbf{Evaluation Setup.} After training on the long-tailed dataset, we evaluate the models on the corresponding balanced test/validation dataset and report top-1 accuracy. To give further insight, we report accuracy on three splits of the set of classes for ImageNet-LT and iNaturalist 2018: Many-shot (>100 images), Medium-shot (20-100 images), and Few-shot (<20 images), adopting from OLTR \cite{openlongtailrecognition}.

\textbf{Hyperparameters for the best performance.}
Backbones and hyper-parameters of our method used for all datasets are listed in Table \ref{hp}. 
\begin{table}[h]
\centering
\begin{tabular}{lccccc}
\hline
Datasets & Epochs & lr (representation/classifier) & Backbone & Init. & s \\
\hline
CIFAR100-LT & 300 & 0.2/0.2 & ResNet-32 & 0. & trainable \cite{kobayashi2021cvpr}\\
CIFAR10-LT  & 300 & 0.2/0.2 & ResNet-32 & 0. & trainable\cite{kobayashi2021cvpr}\\
ImageNet-LT & 90 & 0.4/0.2 & ResNet-10 & -1. & trainable\cite{kobayashi2021cvpr}\\
iNaturalist 2018 & 200 & 0.4/0.2 & ResNet-50 & 1. & 250 \\
\hline
\end{tabular}
\caption{Choice of hyper-parameter in all datasets. lr: Initial learning rate of GSD with cosine scheduler. Init: Initialization of $w_\rho$. Trainable $s$ are implemented following \cite{kobayashi2021cvpr}}
\label{hp}
\end{table}

\subsection{Impact of epoch number.}

As Table \ref{wi_e150} shows, when using a learning rate of 0.2, the overall performance of our method improves with more training epochs, indicating inadequate training. However, we note that such an improvement is attributed to the accuracy improvement of Class "Many". Meanwhile, the accuracy of Class "Few" decreases slightly with more training epochs. It is likely due to the fact that the model weighs more on high-frequency classes with longer training time. Therefore, we increase the learning rate while the same training epoch (Table 2 in \emph{Main text}).

\begin{table}[h]
\centering
\begin{tabular}{lcccccccc}
\hline
& \multicolumn{4}{c}{90 epochs}&\multicolumn{4}{c}{150 epochs}\\
Initialization & Many & Medium & Few & All  & Many & Medium & Few & All \\
\hline
2.0 & 55.3 & 40.4 & 22.9 & 43.7 & 56.9 & 40.6 & 22.1 & 44.3 \\
1.0 & 55.1 & 40.2& 23.0 & 43.7 & 56.9 & 40.6 & 22.0 & 44.3 \\
0 & 55.0 & 40.1 & 22.4 & 43.3& 56.3 & 39.8 & 21.7 & 43.9 \\
-1.0 &55.6 & 40.3 & 22.8 & 43.7 & 57.4 & 40.8 & 21.7 & 44.5 \\
-2.0 & 55.4 & 40.3 & 22.3 & 43.5 & 56.3 & 40.5 & 22.2 & 44.0 \\
\hline
\end{tabular}
\caption{Top 1 accuracy for ImageNet-LT (ResNet-10) with various $w_{\rho}$ initialization (Init.) values. Initial learning rate: 0.2. }
\label{wi_e150}
\end{table}

\subsection{Class-wise $\rho$ optimization}

\begin{figure}[h]
	\centering
	\includegraphics[width=0.4\textwidth]{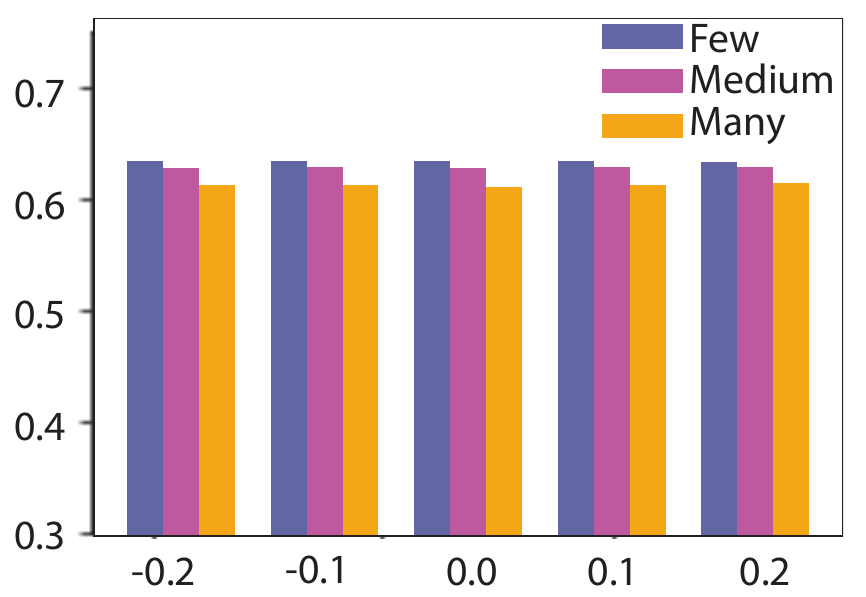}
\caption{Bar graph of $\rho$ values at 90th epoch with respect to different weight initialization values. Three sets of class are plotted include few (<20), medium (20-100) and many (>100). }
	\label{weight_init}
\end{figure}

\textbf{Convergence of $\bm{\rho}$.} Regardless of initialization, $\bm{\rho}$ are able to converge to similar values (Figure \ref{weight_init}), demonstrating our method is robust against initialization.

\begin{table}[h]
\centering
\begin{tabular}{lcccc}
\hline
Sampling method & Many & Medium & Few & All \\
\hline
Class balanced sampling \cite{kang2019decoupling} & 56.2 & 40.9 & 24.1& 44.5\\
Meta-sampling \cite{Ren2020balms} (lr = 0.005) & 54.0 & 42.0 & 23.0 & 44.0\\
Meta-sampling \cite{Ren2020balms} (lr = 0.01) & 53.8 & 41.7 & 25.3 & 44.1\\
Meta-sampling \cite{Ren2020balms} (lr = 0.05) & 52.3 & 41.2 & 27.7 & 43.7\\
\hline
\end{tabular}
\caption{Top 1 accuracy for ImageNet-LT (ResNet-10) with different sampler in classifier learning. We use 3 different learning rates in meta sampling.}
\label{sampler}
\end{table}

\subsection{Impact of the sampler in decoupled training} 

The sampler is demonstrated to be critical when training with an imbalanced dataset, especially in classifier learning. To assess which sampler yields better performance for WCDAS, we compare two predominant sampling approaches: class-balanced sampler and meta sampler. For a fair comparison, we conducted three experiments with a meta sampler using different learning rates. Table \ref{sampler} shows that a class-balanced sampler consistently shows better results than a meta sampler when considering all classes. However, a meta-sampler provides a more balanced accuracy across classes with medium or few examples.  

\subsection{Comparison with selected SOTA methods using same settings: large learning rate.}

Table \ref{tableii} shows the comparison of our method with SOTA softmax-based methods. We note that the results are directly copied from the original paper. \blue{Our method shows superior performance. We also note that those methods show no improvement beyond the results from their original papers when applying a larger learning rate (\emph{Table 4}), indicating that a learning rate of 0.2 is sufficient or optimal for those methods. }

\begin{table}[h]
\centering
\begin{tabularx}{\textwidth}{lXXXX|XXXX}
\hline
Dataset & \multicolumn{4}{c|}{Imagenet-LT} & \multicolumn{4}{c}{iNaturalist 2018} \\
\hline
 & Many & Medium & Few & All & Many & Medium & Few & All \\
\hline
Focal loss \cite{DBLP:journals/corr/abs-1708-02002} & 36.4 & 29.9 & 16.0 & 30.5 & - & - & - & 61.1  \\
OLTR \cite{openlongtailrecognition} & 43.2 & 35.1 & 18.5 & 35.6 & 65.9 & 66.3 & 63.6 & 65.4 \\
Center loss \cite{10.1007/978-3-319-46478-7_31} & 53.1 & 35.0 & 15.6 & 39.2 & 71.5 & 66.0 & 61.8 & 65.8 \\
cRT \cite{kang2019decoupling} & 52.3 & 39.5 & 23.2 & 41.8 & 73.2 & 68.8 & 68.9 & 69.3 \\
LWS \cite{kang2019decoupling} & - & - & - & 41.4 & 71.5 & 71.3 & 69.7 & 70.7\\
LDAM loss \cite{cao2019learning} & - & - & - & 36.1 & - & - & - & 64.6 \\
$\tau$-normalized \cite{kang2019decoupling} & 51.9 & 38.3 & 22.5 & 40.6 & 71.1 & 68.9 & 69.3 & 69.3 \\
BALMS \cite{Ren2020balms} & 50.3 & 39.5 & \textbf{25.3} & 41.8 & - & - & - & -\\
\hline
Angular based Softmax \\
\hline
L-Softmax \cite{liu2016large} & 53.7 & 35.1 & 16.4 & 39.5 & 71.2 & 66.3 & 60.9 & 64.7\\
AM-Softmax \cite{8953658} & 54.0 & 36.0 & 18.6 & 40.5 & 72.5 & 67.6 & 63.2 & 66.4\\
t-vMF Similarity \cite{kobayashi2021cvpr} & 55.2 & 40.6 & 22.3 & 43.7 & 74.2 & 72.1 & 69.9 & 71.1 \\
\hline
WCDAS (class-balanced) & \textbf{56.2} & 40.9 & 24.1 & \textbf{44.5} & \textbf{75.5} & \textbf{72.3} & 69.8 & \textbf{71.8} \\
WCDAS (meta) &  53.8 & \textbf{41.7} & \textbf{25.3} & 44.1 & 71.4 & \textbf{72.3} & \textbf{70.5} & 70.8  \\
\hline
\end{tabularx}
\caption{Top 1 accuracy for ImageNet-LT (ResNet10 \cite{https://doi.org/10.48550/arxiv.1512.03385}) and iNaturalist 2018 (ResNet50 \cite{https://doi.org/10.48550/arxiv.1512.03385}). Results are copied directly from the original papers.}
\label{tableii}
\vspace{-4mm}
\end{table}

\bibliographystyle{plain}  

\bibliography{main}

\end{document}